\documentclass[10pt,twocolumn,letterpaper]{article}
\usepackage[pagenumbers]{cvpr}
\usepackage[T1]{fontenc}
\usepackage{graphicx}
\usepackage{amsmath,amssymb}
\usepackage{booktabs}
\usepackage{multirow}
\usepackage{xspace}
\usepackage{xcolor}
\usepackage{tikz}
\usetikzlibrary{arrows.meta,positioning,fit,backgrounds,calc,shapes.geometric}

\newcommand{\ours}{PostMesh\xspace}
\newcommand{\mf}{MeshFlow\xspace}
\newcommand{\tf}{TriFlow\xspace}
\newcommand{\Vobj}{$V_{\text{obj}}$\xspace}
\newcommand{\rdens}{r_{\text{dens}}}

\usepackage{listings}
\definecolor{cvprblue}{rgb}{0.21,0.49,0.74}
\usepackage[pagebackref,breaklinks,colorlinks,allcolors=cvprblue]{hyperref}
\newcommand\blfootnote[1]{\begingroup\renewcommand\thefootnote{}\footnotetext{#1}\endgroup}
\title{PostMesh: Self-Rewarded Preference Post-Training\\for Flow-Based Artist Mesh Generation}
\author{Gihyun Kwon$^{\dagger}$\\
Amazon\\
{\tt\small ghkwon@amazon.com}}
\begin{document}
\twocolumn[{%
\renewcommand\twocolumn[1][]{#1}%
\maketitle
\vspace{-2.0em}
\begin{center}
\captionsetup{type=figure}
\includegraphics{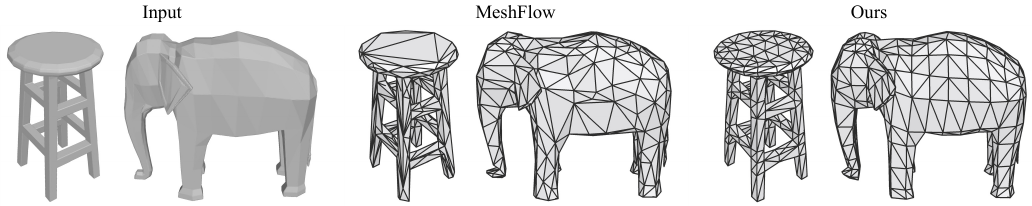}\\[-2pt]
\vspace{-0.6em}
\captionof{figure}{\textbf{Generated artist meshes from \ours.} Our PostMesh post-trains a fast flow-based artist-mesh generator (\mf) on its own samples, without an artist mesh. Two test inputs (shaded), then pretrained \mf and \ours outputs with mesh grids: on these inputs, \ours removes the abnormally long triangles and makes the non-uniform meshes uniform.}
\label{fig:teaser}
\end{center}
\vspace{0.1em}
}]
\blfootnote{$^{\dagger}$Work done outside Amazon.}
\begin{abstract}
Artistic mesh generators turn a dense 3D object into an artist mesh: a compact, quad-dominant mesh that is easy to edit.
Most of current models are auto-regressive methods which decode a mesh token by token, take tens to hundreds of seconds per mesh.
Diffusion- and flow-based generators are much faster, but their meshes still fall short of artist quality.
To our knowledge, \ours is the first post-training method for a diffusion- or flow-based artist mesh generator, and it needs no expensive artist-made ground truth: the generator is trained on its own samples, scored by a reward computed from the dense input and the output.
We propose an artistic mesh reward that keeps the output consistent with the dense input while making the mesh structure into artistic. We also propose a multi-negative flow preference optimisation that transfers the reward to the generator.
After a short LoRA post-training, \ours is comparable to or better than slow autoregressive generators on most artist-mesh quality metrics and in a blind VLM preference test, at a fraction of their time (2.53\,s against 26.98--471.84\,s per mesh).
\end{abstract}

\section{Introduction}
\label{sec:intro}

Artist meshes are compact, quad-dominant and laid out in edge loops, which makes them easy to edit, rig and texture.
Artists build them by hand, so they are hard and expensive to make.

Recently, there has been model-based approaches which automate artistic mesh creation.  PolyGen~\cite{nash2020polygen} generates the vertices and then the faces of a mesh with transformers, and MeshGPT~\cite{siddiqui2024meshgpt} encodes faces as learned tokens and decodes them one at a time.
Later autoregressive (AR) models condition on the input shape and shorten the token sequence: MeshAnything~\cite{chen2025meshanything} turns a point cloud of a shape into an artist mesh, MeshAnything V2~\cite{chen2025meshanythingv2} represents a face by a single vertex token wherever possible, Mesh Silksong~\cite{song2026silksong} layers and orders the vertices so that each is tokenised only once, and FastMesh~\cite{kim2026fastmesh} generates only the vertices autoregressively and predicts all faces in one pass.
The above methods follows auto-regressive (AR) approach which decodes one token at a time. Although these methods can generate high-quality artist meshes, but generating one token at a time makes them expensive and slow.

\emph{Diffusion- and flow-based} generators instead produce the whole mesh in parallel. PolyDiff~\cite{alliegro2023polydiff} denoises a quantised triangle soup by discrete diffusion, \tf~\cite{li2026triflow} predicts nearest-vertex vector fields by latent flow matching, and \mf~\cite{li2026meshflow} compresses an artist mesh, its vertices and connectivity, into a compact per-vertex latent with a MeshVAE. In \mf, a rectified-flow transformer conditioned on the input shape generates all latent tokens at once, and the MeshVAE decoder turns them into vertices and faces.
Because the whole latent is generated in a fixed number of denoising steps instead of token by token, these models are much faster than auto-regressive approaches.
Despite their speed, their meshes still fall short of the best AR models, with long, thin triangles (\cref{fig:teaser}) and fragmented parts, so there is room for improvement. These defects live in the decoded mesh, which the flow-matching loss never sees.

For AR models, much of the recent progress came from post-training with preference optimisation or reinforcement learning (RL) on the token likelihood: DeepMesh~\cite{zhao2025deepmesh} and Mesh-RFT~\cite{liu2025meshrft} use Direct Preference Optimization, QuadGPT~\cite{liu2026quadgpt} use tDPO and Mesh-Pro~\cite{zhou2026meshpro} use online RL-based training.
Reward and preference optimisation have also been applied to dense 3D generators~\cite{ye2024dreamreward,li2025dso,zhou2026dho,ning2026flow3dopd}, whose output is an implicit field or an iso-surface. 
However, to our knowledge, no diffusion- or flow-based artist-mesh generator has been post-trained in this way.
Collecting more artist meshes is expensive, and training the flow model on its own best samples helps less: imitating them (RAFT~\cite{dong2023raft}) made outputs \emph{more} fragmented in our study (\cref{tab:ablation_main}), and in our ablations, fine-tuning on the reward-selected samples with the flow-matching loss alone loses most of the quad-structure gain of preference optimisation (\cref{sec:ablations}).
We therefore post-train the generator without ground truth artist meshes, delivering a reward to it directly through preference optimisation.

\ours is the first post-training method for a diffusion- or flow-based artist-mesh generator, and its \emph{artistic mesh reward} is the first reward for such a generator that keeps the output consistent with the dense input while steering it toward the structure of artist meshes, without any artist mesh.
For training, we propose a \emph{gating algorithm} which also rejects samples with more flipped faces than their group, turns each group of samples into one winning and several weighted losing samples. Also we use  \emph{multi-negative flow preference optimisation}, Diffusion-DPO~\cite{wallace2024diffusiondpo} with several losing samples and an RPO likelihood term~\cite{liu2024rpo}, transfers the reward to the generator.
Our contributions are:
\begin{itemize}[leftmargin=*,itemsep=1pt,topsep=2pt]
\item To our knowledge, the \textbf{first} post-training of a diffusion- or flow-based artist-mesh generator, done by self-training on its own samples without any artist mesh.
\item We propose an artistic mesh reward, to our knowledge the \textbf{first} for guiding the generated mesh to be consistent with the dense input and follows artist-mesh structure: silhouette IoU and depth agreement on renders to keep the shape consistency, and quad strips and tessellation, slivers, vertex valence, tiny and stray components and a face-count bound reward for artist-preferred mesh generation.
\item A gating algorithm and a multi-negative flow preference optimisation that turn the reward into weighted winning and losing samples and transfer it to a flow model.
\end{itemize}
Overall, \ours is better than or comparable to slow autoregressive generators on most artist-mesh quality metrics and in a blind VLM preference test, and it is far more efficient in time.

\section{Related Work}
\label{sec:related}

\paragraph{Autoregressive artist-mesh generation.}
Beyond the models above, AR work has mostly restructured the token sequence: MeshXL~\cite{chen2024meshxl} trains large transformers on coordinate sequences, BPT~\cite{weng2025bpt} shortens them by block-wise indexing and patches of faces, EdgeRunner~\cite{tang2025edgerunner} compresses meshes into shorter token sequences, Meshtron~\cite{hao2024meshtron} trains an hourglass transformer on truncated sequences to reach much larger meshes, and QuadGPT~\cite{liu2026quadgpt} generates quads natively. All decode sequentially, so cost grows with face count.

\paragraph{Diffusion and flow artist-mesh generation.}
Besides PolyDiff, \tf and \mf, MeshCraft~\cite{he2025meshcraft} runs a flow transformer over VAE face latents for a chosen face count, PolyFlow~\cite{wang2026polyflow} denoises continuous per-vertex embeddings from which the connectivity is recovered, and LATO~\cite{zhao2026lato} encodes vertices and connectivity in a joint sparse-voxel latent, which LATO.2~\cite{long2026lato2} factorises into a vertex flow followed by a connectivity flow.
Our base model is the \mf of Li~\etal, unrelated to the equivariant MeshFlow of Sun~\etal~\cite{sun2026meshflowequi}. To our knowledge, none of these reports an RL or preference post-training stage.

\paragraph{Post-training generative models.}
DPO~\cite{rafailov2023dpo} optimises pairwise preferences without a reward model, and RPO~\cite{liu2024rpo,pang2024irpo} adds a likelihood term on the preferred sample. Diffusion-DPO~\cite{wallace2024diffusiondpo} uses the denoising error of a re-noised sample as the likelihood, Flow-GRPO~\cite{liu2025flowgrpo} turns the ODE sampler of a flow model into an SDE to apply GRPO, and RAFT~\cite{dong2023raft} fine-tunes on reward-ranked samples.
In 3D, DreamReward~\cite{ye2024dreamreward} learns a human-preference reward for text-to-3D, DSO~\cite{li2025dso} aligns a 3D generator with feedback from a physics simulator, and concurrent DHO~\cite{zhou2026dho} (online RL) and Flow3D-OPD~\cite{ning2026flow3dopd} (distillation from DPO teachers) post-train flow-matching generators of iso-surfaced dense geometry.
Artist-mesh post-training so far is autoregressive and acts on token likelihoods (\cref{sec:intro}); Mesh-RFT~\cite{liu2025meshrft}, for instance, uses face-masked DPO with a topology score.
Our policy instead is a flow model with no token likelihood, behind a frozen, non-differentiable decoder. To our knowledge, \ours is the first to post-train a diffusion- or flow-based artist-mesh generator.\label{sec:related-end}%

\begin{figure*}[t]
\centering
\includegraphics[width=\textwidth]{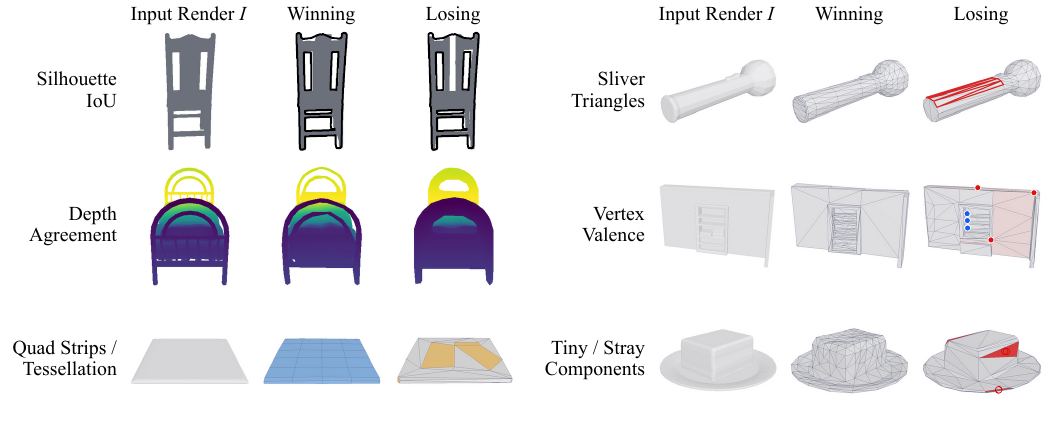}
\caption{\textbf{Artistic Mesh Rewards.} Each row shows the dense input $I$ and the winning and a losing sample of the same input. In each row the named term favours the winning sample. Silhouette IoU and depth agreement compare renders of the output with renders of $I$: they penalise missing or extra parts (filled silhouettes with the outline of $I$) and surfaces pushed in or out (depth maps). Quad strips and tessellation reward quads laid out in rings and strips. The sliver term penalises needle-like triangles (red edges). Vertex valence, the number of edges at a vertex, penalises poles and fans: a vertex with 10 or more edges (red dots) is typically the hub of a fan of thin triangles, and a vertex with at most 2 edges (blue dots) belongs to a single triangle, a loose corner. Both break the edge loops of a regular layout. Tiny and stray components penalises specks and floating pieces (red).}
\label{fig:anatomy}
\vspace{-0.8em}
\end{figure*}

\section{Method}
\label{sec:method}

\begin{figure}[t]
\centering
\includegraphics[width=\columnwidth]{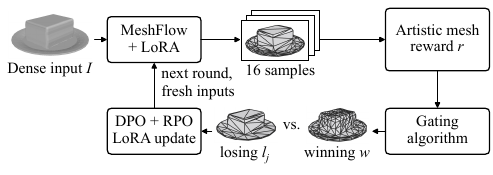}
\caption{\textbf{Training Overview.} For each dense input $I$, the current model (\mf with LoRA) draws a group of samples then the artistic mesh reward scores them. The gating algorithm keeps a winning sample $w$ and weighted losing samples $l_j$. With those samples, a DPO and RPO step updates only the LoRA adapters, and the next round samples fresh inputs with the updated model. }
\label{fig:pipeline}
\end{figure}
\subsection{Problem and Overview}
\label{sec:prelim}
\mf~\cite{li2026meshflow} encodes an artist mesh as a per-vertex latent $z_0$. A rectified-flow transformer $v_\theta(z_\sigma,\sigma,c)$ predicts $\epsilon-z_0$ along $z_\sigma=(1-\sigma)z_0+\sigma\epsilon$, conditioned on surface samples $c$ of a dense input mesh $I$. A frozen decoder turns $z_0$ into vertices, edges and 3-clique faces, so topology defects appear only after decoding, out of sight of the flow-matching loss.
We improve the model by \emph{self-training}: it learns only from its own samples, and no artist mesh is added to the training data, the rewards or the selection of winning and losing samples. We train LoRA adapters~\cite{hu2022lora} upon pretrained flow model.
\paragraph{Training pipeline.}
For each dense input, the model draws a group of samples (16 in our setting). Each sample is decoded and scored by the artistic mesh reward. A gating algorithm then picks the best plausible sample as the winner and turns clearly worse or implausible samples into weighted losing samples, and the preference loss raises the likelihood of the winner under the model relative to the losers (\cref{fig:pipeline}). Each round repeats this on fresh inputs with the updated model.

\subsection{Artistic Mesh Reward}
\label{sec:reward}
An artist mesh should (i) follow the shape of the dense input and (ii) have the structure artists need: quads laid out in edge loops, no needle-like triangles, no poles, no specks or stray pieces, and a sensible face budget. Since we do not have artist mesh for training, we score (i) against renders of the input and (ii) on the output alone.
\Cref{fig:anatomy} shows each term on real training pairs.
For the two input-consistency terms we render the output $O$ and $I$ from fixed views in the frame of $I$. Per view, $S_O,S_I$ are the silhouettes and $d_O,d_I$ the depths, and scores are averaged over views.

\subsubsection{Silhouette IoU}
\label{sec:r-sil}
$t_{\text{sil}}$ scores the intersection over union of the silhouettes,
\begin{equation}
\mathrm{IoU}=|S_O\cap S_I|\,/\,|S_O\cup S_I| .
\label{eq:sil}
\end{equation}
Without an artist mesh, the outline of the dense input is the most direct evidence of what the output must contain. A missing part, such as a chair leg or a handle, or an extra blob changes the outline in some view and lowers the IoU. Although it makes strong shape consistency between dense inputs and outputs, there is clear limit that it sees nothing inside the outline, such as offsets or flipped faces, which is why additional reward is required to complements it.

\subsubsection{Depth Agreement}
\label{sec:r-depth}
$t_{\text{depth}}$ scores the mean depth difference  where both silhouettes cover a pixel, in units of the bounding-box diagonal,
\begin{equation}
\mathrm{depth}_{L1}=\tfrac{1}{|S_O\cap S_I|}\textstyle\sum_{p\in S_O\cap S_I}|d_O(p)-d_I(p)| ,
\label{eq:depth}
\end{equation} 
where the $p$ is render from generated mesh within silhouette intersection.
Two meshes can share every outline while one has surfaces pushed in or out: a sunken seat, a bulging side, a part of the wrong thickness. The depth maps reveal these offsets. Together with silhouette IoU, depth agreement forms the shape constraint rewards.  Because the two terms compares renders, not surfaces, they do not prefer any triangulation or mesh structure. 

\subsubsection{Quad Strips \& Tessellation}
\label{sec:r-topo}
Artists build meshes from quads arranged in edge loops, which makes selection, UV unwrapping, subdivision and deformation predictable. Both terms first merge adjacent triangle pairs into quads: a pair merges if its two triangles are within $10^\circ$ of coplanar and their union is a convex quad with all corners in $[45^\circ,135^\circ]$, greedily in decreasing order of the smallest corner of the quad. Let $\mathcal{Q}$ be the merged quads and $\mathcal{T}$ the triangles left over.

\noindent\textbf{Quad strips.} Stepping from a quad across one of its sides into the neighbouring quad, and on across the opposite side, traces a strip; it ends at a triangle, a boundary or a non-manifold edge, and a strip that closes on itself is a ring. Each quad lies on two strips, one per pair of opposite sides. With $\Pi$ the set of strips, $t_{\text{QS}}$ scores the quad score
\begin{equation}
\mathrm{QS}=\tfrac{1}{|\mathcal{Q}|}\bigl|\{q\in\pi:\ \pi\in\Pi\ \text{a ring or}\ |\pi|\ge4\}\bigr| ,
\label{eq:qs}
\end{equation}
the share of quads on a ring or on a strip of at least four quads, inspired by the quad rings and lines of QuadGPT~\cite{liu2026quadgpt} and Mesh-Pro~\cite{zhou2026meshpro}. It rewards quads that continue into edge loops, so a mesh can have many quads and still a low QS if they are scattered among triangles.

\noindent\textbf{Tessellation.} $t_{\text{TS}}$ scores the Mesh-RFT topology score~\cite{liu2025meshrft},
\begin{equation}
\mathrm{TS}=0.4\,\tau_1+0.2\,\tau_2+0.3\,\tau_3+0.1\,\tau_4 ,
\label{eq:ts}
\end{equation}
with the quad ratio $\tau_1=|\mathcal{Q}|/(|\mathcal{Q}|+|\mathcal{T}|)$, the right-angle score $\tau_2=1-\tfrac{1}{|\mathcal{Q}|}\sum_{q}\sum_{k=1}^{4}|\theta_{q,k}-90^\circ|/360^\circ$ over the corners $\theta_{q,k}$ of each quad, a score for equal opposite sides and side lengths ($\tau_3$) and one for similar shapes of neighbouring quads ($\tau_4$). Exact definitions are in \cref{supp:metrics}. 

TS thus rewards quads that are well shaped where they are (limited on neighboring mesh connectivity) , and QS measures how well quads continue into strips and rings (larger set mesh connectivity check). Both terms are blind to shape, which is why they are paired with silhouette IoU and depth agreement.

\begin{figure*}[t]
\centering
\includegraphics[width=\textwidth]{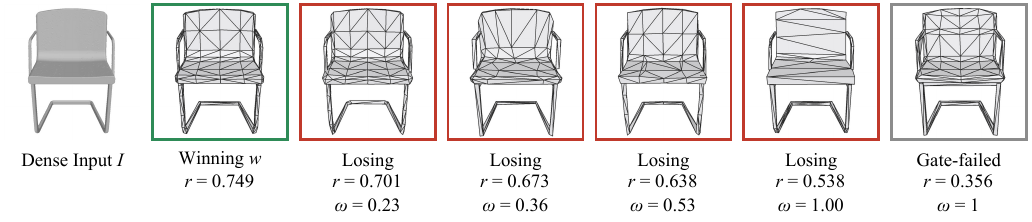}
\caption{\textbf{Training Group Examples.} The dense input $I$ and six of its 16 samples are shown. Green frame: the winning sample $w$. Red: gate-passing losing samples with their preference weight $\omega$. Grey: a sample that failed a gate, which counts as a losing sample with $\omega=1$. $r$: artistic mesh reward. }
\label{fig:group}
\end{figure*}
\subsubsection{Sliver Triangles}
\label{sec:r-sliver}
$t_{\text{sliver}}$ scores the area share of needle-like triangles that were not merged into quads,
\begin{equation}
s_{\text{sliver}}=\textstyle\sum_{f\in\mathcal{T},\,\alpha_f<5^\circ}A_f\,\big/\sum_{f\in\mathcal{F}}A_f ,
\label{eq:sliver}
\end{equation}
with $\mathcal{F}$ all triangles of $O$, $A_f$ the area and $\alpha_f$ the smallest corner of triangle $f$.
Such needle-like triangles shade badly, cannot become part of a quad layout, and make editing and subdivision unstable; \cref{fig:teaser} shows them on outputs of the pretrained model. The term measures sliver area rather than a count, so long slivers that run across a surface cost more than a few tiny ones.

\subsubsection{Vertex Valence}
\label{sec:r-val}
$t_{\text{val}}$ scores the share of vertices with extreme valence $\deg(v)$, the number of edges at $v$,
\begin{equation}
s_{\text{val}}=\bigl|\{v\in\mathcal{V}:\deg(v)\le2\ \text{or}\ \deg(v)\ge10\}\bigr|\,\big/\,|\mathcal{V}| ,
\label{eq:val}
\end{equation}
with $\mathcal{V}$ the vertices that have at least one edge.
A vertex with 10 or more edges is a pole, typically the hub of a fan of thin triangles. A vertex with at most 2 edges belongs to a single triangle, a loose corner. Both interrupt edge loops, pinch the surface under subdivision and deform badly when the mesh is animated. Artists do place some poles, for example at corners, so only extreme valences are penalised.

\subsubsection{Tiny \& Stray Components}
\label{sec:r-valid}
$t_{\text{frag-n}}$ scores $\log(1{+}n_{\text{tiny}})$, with $n_{\text{tiny}}$ the number of components with at most 4 faces, and $t_{\text{frag-stray}}$ scores $s_{\text{stray}}$, the area share of stray components, those that cover less than 1\% of the surface and have a boundary edge or at most 4 faces. Together they prevent specks and floating pieces, which an artist must find and delete by hand and which renders show only as hairline cracks. The two terms complement each other: the count reacts to many specks even when their area is negligible, and the area reacts to a few larger stray pieces. Fusing real parts is not rewarded: artist meshes are multi-part, and the gating algorithm rejects samples with far fewer real parts than their group.

\subsubsection{Face Count Bound}
\label{sec:r-dens}
With $F$ the face count of the output and $F_t$ a target predicted from the input,
\begin{equation}
r_{fc}=\mathrm{clip}\bigl(1-|\log_2(F/F_t)|/b,\,0,\,1\bigr),
\label{eq:dens}
\end{equation}
which is 1 at the target and falls on both sides within a band of width $b$. Because the score is on a log scale, twice the target face count costs as much as half of it. It keeps the face budget sensible: it stops the policy from shedding faces, which training did without it, and from copying the dense input.

\paragraph{Overall reward.}
Each term except $\rdens$ maps its raw score linearly to $[0,1]$ between two fixed anchors. The reward is the weighted sum of all terms,
\begin{equation}
\begin{aligned}
r={}&\lambda_{\text{sil}}t_{\text{sil}}+\lambda_{\text{depth}}t_{\text{depth}}+\lambda_{\text{QS}}t_{\text{QS}}+\lambda_{\text{TS}}t_{\text{TS}}\\
&+\lambda_{\text{sliver}}t_{\text{sliver}}+\lambda_{\text{val}}t_{\text{val}}+\lambda_{\text{tiny}}t_{\text{frag-n}}\\
&+\lambda_{\text{stray}}t_{\text{frag-stray}}+\lambda_{F}r_{fc} ,
\end{aligned}
\label{eq:reward}
\end{equation}
with weights $\lambda$ that sum to 1 (values in \cref{supp:reward}).

\subsection{Gating Algorithm}
\label{sec:pairs}
The gating algorithm turns a group of scored samples into preferences. Its rules compare each sample with the other samples of the same input, so they adapt to how hard that input is.
A sample fails a gate if it is implausible relative to its group: its Chamfer or Hausdorff distance to $I$ is far above the group median, it has more flipped faces or more tiny components than the group, it has far fewer real parts, or it fails to decode.
The winning sample $w$ is the best gate-passing sample under $r$. The losing samples are every gate-failed sample, with weight 1 whatever its reward, and every gate-passing sample clearly below $w$, by more than half a group standard deviation.
Finally, a pair is dropped if $w$ is worse than the losing sample in fidelity to $I$, coverage of $I$, face count or centring, so that no preference teaches the model to trade these away.
\Cref{fig:group} shows a real group of training samples. Detailed thresholds and gating alrogithm criteria are in \cref{supp:gating}.

\begin{table*}[t]
\centering
\footnotesize
\setlength{\tabcolsep}{0pt}
\begin{tabular*}{\textwidth}{@{\extracolsep{\fill}}lrrrrrrrrrrrrr@{}}
\toprule
Method & TS$\uparrow$ & QS$\uparrow$ & DFF$\downarrow$ & Frag.$\downarrow$ & CD$\downarrow$ & HD$\downarrow$ & ECD$\downarrow$ & SI$\downarrow$& NME$\downarrow$& UV$_\Delta$ $\downarrow$& Faces & s/mesh$\downarrow$ & VLM pref.\\
\midrule
\multicolumn{11}{@{}l}{\emph{Autoregressive}}\\
FastMesh V4K~\cite{kim2026fastmesh} & 0.559 & 0.035 & 0.990 & 1.000 & 3.247 & 8.315 & 5.736 & 35.33&79.43&450.53& 12490 & 26.98 & 92.5\%\\
MeshAnything~\cite{chen2025meshanything} & 0.746 & 0.756 & 0.319 & 0.317 & 5.055 & 15.624 & 9.823 &11.07&2.88&166.74& 439 & 68.73 & 82.5\%\\
MeshAnything V2~\cite{chen2025meshanythingv2} & 0.782 & 0.776 & 0.335 & 0.620 & 4.704 & 14.496 & 9.556 &12.35&4.63&121.55& 1045 & 73.77 & 72.5\%\\
Mesh Silksong~\cite{song2026silksong} & 0.708 & 0.790 & 0.090 & 0.533 & 2.981 & 9.735 & 5.654 &2.52&1.88&63.55& 2344 & 471.84 & 62.5\%\\
\midrule
\multicolumn{11}{@{}l}{\emph{Flow-based}}\\
\mf~\cite{li2026meshflow} & 0.575 & 0.466 & 0.206 & 0.743 & 2.760 & 6.169 & 8.149 &3.43&11.04&84.26& 609 & 2.17 & 80.0\%\\
\textbf{\ours} (\mf + ours) & 0.706 & 0.804 & 0.121 & 0.647 & 2.697 & 6.089 & 8.123 &1.50&4.78&70.08& 728 & 2.53 & --\\
\bottomrule
\end{tabular*}
\caption{\textbf{Quantitative Scores.} \ours has the highest QS, and its TS is on par with Mesh Silksong and close to the two MeshAnything models. Its DFF is lower than that of every baseline except Mesh Silksong, which it approaches, and its fragmentation is lower than that of \mf, though above most autoregressive baselines. Its CD and HD are the lowest. Its ECD is higher than that of Mesh Silksong and FastMesh, likely because they use several times more faces. Above all, \ours runs almost as fast as \mf (2.53 vs.\ 2.17\,s per mesh), one to two orders of magnitude faster than the autoregressive models, while matching or exceeding them on most metrics. VLM pref.: blind VLM preference rate for \ours against each method, on 60 objects independent of the 300 test objects. The judge prefers \ours over every method, and slightly over Mesh Silksong.}
\label{tab:main}
\vspace{-0.6em}
\end{table*}
\subsection{Preference Loss}
\label{sec:dpo}
Each kept pair compares $w$ with a losing sample $l_j$ of the same input.

\subsubsection{Flow-Matching Error as a Likelihood Proxy}
\label{sec:loss-fm}
The loss should make the model more likely to produce $w$ than $l_j$, but a flow model gives no cheap, differentiable likelihood of a finished mesh. As in Diffusion-DPO~\cite{wallace2024diffusiondpo}, we re-noise the latent $z_0$ of the finished sample and measure how well the model denoises it; a lower error means the model finds the sample more likely. The error, averaged over latent entries, stands in for the negative log-likelihood:
\begin{equation}
\ell_\theta(z_0)=\tfrac{1}{|z_0|}\bigl\|v_\theta(z_\sigma,\sigma,c)-(\epsilon-z_0)\bigr\|^2 ,
\label{eq:fm}
\end{equation}
with one $(\epsilon,\sigma)$ per input shared by $w$ and its losing samples, so that differences come from the samples, not the noise.

\subsubsection{Margin Against the Frozen Reference}
\label{sec:loss-margin}
The margin asks one question: has fine-tuning made the model better at producing $w$ than at producing $l_j$? Each error is measured against the same error $\ell_{\text{ref}}$ of the frozen pretrained \mf ,
\begin{equation}
d_j=\bigl(\ell_\theta(w)-\ell_{\text{ref}}(w)\bigr)-\bigl(\ell_\theta(l_j)-\ell_{\text{ref}}(l_j)\bigr).
\label{eq:margin}
\end{equation}
$d_j<0$ when the policy improved its fit to $w$ more than its fit to $l_j$. Because only the genuine change is considered during training, a sample that was easy for the model from the start does not contribute to the training.

\subsubsection{Multi-Negative Weighting}
\label{sec:loss-weight}
Instead of one pair per input, every losing sample of the group contributes, and its weight $\omega_j$ says how bad it is:
\begin{equation}
\mathcal{L}_{\text{DPO}}=\textstyle\sum_j\omega_j\,\phi(d_j)\big/\sum_j\omega_j ,
\label{eq:ldpo}
\end{equation}
with $\phi$ a soft loss on the preference margin (defined below). A gate-passing losing sample weighs its reward gap to $w$ divided by the largest gap in the group, so a clearly worse sample pushes harder than a near-tie. A gate-failed one weighs $\omega_j=1$, because failing a gate is not a matter of degree. Using several losing samples draws more signal from each group than a single pair; when an input has many, each update uses a random subset, redrawn every update. 

\subsubsection{The Scale \texorpdfstring{$\beta$}{β} and the Log-Sigmoid}
\label{sec:loss-beta}
Each losing sample contributes $\phi(d)=\mathrm{softplus}(\tfrac{\beta}{2}d)=-\log\mathrm{sigmoid}(-\tfrac{\beta}{2}d)$.
This turns the margin into a soft classification loss: it costs little once $w$ is clearly preferred and grows linearly when $l_j$ is preferred, and $\beta$ sets how sharp that switch is (value in \cref{supp:loss}).

\subsubsection{RPO Term on the Winning Sample}
\label{sec:loss-rpo}
The DPO term only looks at the gap between $w$ and $l_j$, so it can be satisfied by letting both errors rise, as long as the error of the losers rises more. The RPO term~\cite{liu2024rpo,pang2024irpo} keeps the winner itself likely by penalising $\ell_\theta(w)$ directly. The per-input loss, averaged over the inputs of an update, with RPO weight $\gamma$, is
\begin{equation}
\mathcal{L}=\mathcal{L}_{\text{DPO}}+\gamma\,\ell_\theta(w).
\label{eq:loss}
\end{equation}
Without the RPO term (DPO only), one training run gave an inconclusive result, so its benefit is not established.

\section{Experiments}
\label{sec:exp}

\subsection{Setup}
\label{sec:setup}
\paragraph{Training.}
Training objects come from Objaverse~\cite{deitke2023objaverse} with at most 100k faces. Training inputs come from a 2,000-object pool disjoint from evaluation. We run two rounds of self-training, updating only the LoRA adapters. At each training round, we use 1000 training iterations with 1000 object data. After one training round we use new batch of 1,000 inputs and apply same training protocol to the previous round. More detailes are in \cref{supp:impl}. 

\paragraph{Test Data.} For testing and evaluation of the post-trained model, we chose another batch of 300 dense objects randomly collected from Objaverse with proper face count band. For ablation study, we selected another independent batch of 200 objects from Objaverse dataset. The test objects have corresponding artist meshes.
\begin{figure*}[t]
\centering
\includegraphics[width=0.9\textwidth]{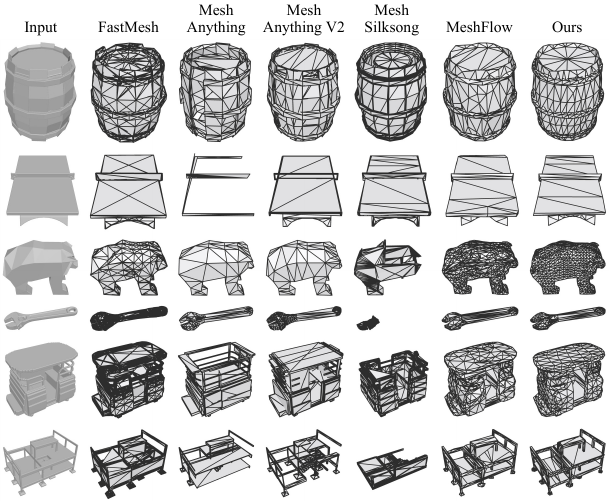}
\caption{\textbf{Qualitative Results.} In this figure, we show the input dense mesh and the generated artist mesh outputs from autoregressive baselines and flow-based models. For the outputs from baselines, they  are broken or incomplete, irregular, or filled with very small faces, which makes them hard for an artist to use, while the \ours outputs are complete and more evenly laid out. The output shows our proposed model have superior performance compared to baselines.}
\label{fig:hero}
\end{figure*}
\paragraph{Metrics.}
\Cref{tab:main} measures fidelity against the artist mesh, topology after the quad merge of \cref{sec:r-topo}. For details, see \cref{supp:metrics}.
\begin{itemize}[leftmargin=*,itemsep=0pt,topsep=1pt]
\item \textbf{CD, HD} ($\times100$): Chamfer and Hausdorff distance between points sampled on the output and on the artist mesh, the average and the worst-case shape error.
\item \textbf{ECD}~\cite{chen2021nmc}: Chamfer distance between points sampled on sharp edges of the two meshes, \ie, how well creases are kept.
\item \textbf{QS}: share of merged quads on a ring or a strip of ${\ge}4$ quads, \ie, continuous edge flow (\cref{eq:qs}).
\item \textbf{TS}~\cite{liu2025meshrft}: Mesh-RFT score of quad share, right angles, even sides and like neighbours (\cref{eq:ts}).
\item \textbf{DFF}: area share of defect faces (on a boundary or non-manifold edge, self-intersecting, degenerate or in a stray fragment), \ie, what an artist must repair.
\item \textbf{Frag.}: share of outputs with at least 5 more small components (under 1\% of the area) than the artist mesh.
\item \textbf{SI}($\times$100): share of faces that self-intersect after two levels of Loop subdivision.
\item\textbf{NME}($\times$100): share of edges shared by more than two faces, which half-edge structures and many tools do not accept.
\item\textbf{UV$_\Delta$}: absolute difference to the artist mesh in UV charts per 1k faces after the same automatic unwrap.
\item \textbf{Faces}: mean face count; \textbf{s/mesh}: seconds per mesh on one L40S GPU.

\end{itemize}

\paragraph{Baselines.}
We compare with autoregressive artist-mesh generators that take a dense mesh: MeshAnything~\cite{chen2025meshanything}, MeshAnything V2~\cite{chen2025meshanythingv2}, Mesh Silksong~\cite{song2026silksong} and FastMesh~\cite{kim2026fastmesh}, which decodes vertices autoregressively and faces in one pass. \mf~\cite{li2026meshflow}, our base model, is the only flow-based one.

\subsection{Quantitative Results}
\label{sec:main}
\Cref{tab:main} compares all methods on the 300 test objects.
Against the autoregressive generators, \ours has the lowest mean CD and HD of all methods. FastMesh and Mesh Silksong have higher CD and HD than both \mf and \ours, which indicates that our outputs keep the shape of the dense input well. For ECD, Mesh Silksong and FastMesh are better than \mf and \ours. We conjecture that this is because they generate several times more faces (2344 and 12490 against 728 for \ours), which makes sharp edges easier to keep. Against the two MeshAnything models, whose face counts are close to ours, \mf and \ours are clearly better on all three shape-consistency metrics (CD, HD and ECD). 

On mesh topology measurement, \ours has the highest QS of all methods, and its TS is on par with that of Mesh Silksong, above that of FastMesh and slightly below that of both MeshAnything models. Compared with our main baseline MeshFlow, our Post-trained model shows meaningful improvements  in both of QS and TS scores.

In terms of defective mesh scores, \ours has a lower DFF than FastMesh and both MeshAnything models, but slightly higher one than Mesh Silksong. For fragmentation score, ours has similar score to MeshAnything v2, but worse than MeshAnything and Mesh Silksong. However, similar to our topology score cases, our post-trained model can clearly reduce the defective meshes compared to our main baseline MeshFlow.

Over its base model, \ours improves QS, TS, DFF and fragmentation at slightly lower CD and HD, with more faces. 
Because QS, TS and DFF partly overlap with our reward terms, we also measure, on the same outputs, properties that no reward term computes: robustness to Loop subdivision, non-manifold edges and automatic UV unwrapping. \ours improves over \mf overall on these as well: after subdivision it has about half the flipped and self-intersecting faces, it has less than half the non-manifold edges, and its UV unwrapping is slightly better. More experiments are in \cref{supp:extra}.

The most important practical difference is speed: \ours keeps the cost per mesh of \mf, below that of the autoregressive models.

\subsection{Blind VLM Preference Test}
\label{sec:perceptual}
QS and TS are reward terms, so better scores on them could reflect the reward rather than better meshes, and a check that does not use them is needed. A study with professional artists would be the natural check but is too costly at this scale. We therefore gave a strong vision-language model, Claude Opus 5.5, explicit artist-mesh criteria and asked it to judge blind. Prompted as a senior 3D artist reviewing meshes for production, it weighs edge flow and topology, triangle quality (few slivers, a sensible density), cleanliness (no floating fragments, holes or flipped faces) and fidelity to the input shape.

In a two-alternative forced choice, the judge saw renders of the dense input and two candidates labelled only A and B, in both orders, on 60 objects separate from the test set. 
The judge preferred \ours over \mf, FastMesh and both MeshAnything models, and slightly preferred it over Mesh Silksong (62.5\%, \cref{tab:main}). For detailed experiment protocol, see \cref{supp:vlm}.

\subsection{Qualitative Results}
\label{sec:qual}
In \Cref{fig:hero}, we show the actual generated meshes from baseline and our proposed model.

For earlier baseline of FastMesh, it suffer from highly fragmented meshes where each face is very tiny and over segmented. In MeshAnything groups, the generated meshes follow the shape of dense inputs while making the meshes into artist-like, but some of the meshes are incomplete and still lots of meshes are thin sliver triangles. In the most recent baseline of Mesh Silksong, it shows high quality artist meshes with proper mesh structure and consistency with input dense mesh, but some cases are suffer from incomplete mesh generation. 

\mf follows the input shapes but lays long, thin triangles across surfaces such as the keg.
\ours keeps the shapes complete and gives a more regular layout on most rows, without fragmented or irregular shaped meshes.

Some autoregressive outputs are incomplete because decoding hit the token limit: 51 of the 300 Mesh Silksong outputs (17\%) reached its 10240-token cap, and 82 of the 300 MeshAnything V2 outputs (27\%) stopped at its token limit. 

\begin{table}[t]
\centering
\small
\setlength{\tabcolsep}{3pt}
\begin{tabular}{@{}lrrrr@{}}
\toprule
 & QS$\uparrow$ & TS$\uparrow$ & DFF$\downarrow$ & Frag.$\downarrow$\\
\midrule
Full setting & 0.746 & 0.672 & 0.111 & 0.663\\
\midrule
\multicolumn{5}{@{}l}{\emph{Reward terms removed} (change vs.\ full setting)}\\
$-$tess./sliv./val. & $-$0.201 & $-$0.066 & $+$0.033 & $-$0.045\\
$-$strips/tess. & $-$0.142 & $-$0.044 & $+$0.018 & $+$0.047\\
$-$silhouette & $-$0.055 & $-$0.020 & $-$0.008 & $-$0.025\\
$-$depth & $-$0.097 & $-$0.033 & $+$0.013 & $+$0.035\\
$-$tiny/stray & $-$0.045 & $-$0.011 & $+$0.039 & $+$0.017\\
\midrule
\multicolumn{5}{@{}l}{\emph{Other post-training objectives} (change vs.\ full setting)}\\
SFT only & $-$0.206 & $-$0.055 & $+$0.091 & $+$0.137\\
RAFT~\cite{dong2023raft} & $-$0.323 & $-$0.099 & $+$0.182 & $+$0.205\\

Flow-GRPO~\cite{liu2025flowgrpo} & $-$0.298 & $-$0.095 & $+$0.102 & $+$0.145\\
\bottomrule
\end{tabular}
\vspace{-0.4em}
\caption{\textbf{Ablation study} on 200 objects separate from the test set. The first row gives the absolute scores of the full setting; every other row gives the change relative to it. Removing the structural reward terms causes the largest drops in QS and TS. Removing a consistency term (silhouette IoU or depth agreement) also lowers QS and TS, while slightly improving consistency scores such as CD (see the appendix).}
\label{tab:ablation_main}
\vspace{-1.2em}
\end{table}

\subsection{Ablation Study}
\label{sec:ablations}
\Cref{tab:ablation_main} removes one component at a time from the full setting and retrains the model on the same candidates, scoring on 200 testing objects.
Among the reward terms, removing the quad-strip and tessellation terms costs the most QS and TS, and removing them together with the sliver and valence terms costs even more. Removing silhouette IoU or depth agreement also lowers QS and TS. Removing the tiny and stray terms lowers QS and raises DFF, which indicates that stray term is related to fragmentation defects.
Fine-tuning on the winning samples without the preference objective (SFT only) loses most of the QS gain and raises DFF and fragmentation, so the preference loss, not the selection alone, carries the improvement.
RAFT and Flow-GRPO stay far below the full setting, with 0.32 and 0.30 lower QS and higher DFF and fragmentation.

\section{Limitations and Conclusion}
\label{sec:conclusion}

\paragraph{Limitations.}
Failure cases are shown in \cref{fig:supp-failure}.
There is no single standard for an artistic mesh, and artists differ in what they want, so our reward is not a golden standard for everyone. The evaluation of artist mesh is subjective, therefore our automatic metrics and VLM study is not a perfect measurement of artist mesh quality. However, the value of \ours lies in the method: it firstly shows how to post-train a fast flow-based generator toward a chosen standard without artist meshes.

\paragraph{Conclusion.}
\ours post-trains a fast flow-based artist-mesh generator on its own samples, with an artistic mesh reward and gated multi-negative preference optimisation, and without artist meshes. It is comparable to or better than slow autoregressive generators on most quality measures and in a blind VLM test while keeping the fast speed of flow-based artist mesh generation model. \label{sec:body-end}%

\clearpage
\appendix
{\noindent\Large \textbf{Supplementary Material}}
\paragraph{Contents.}
This appendix gives the implementation details (\cref{supp:impl}), the gating algorithm (\cref{supp:gating}), the reward constants (\cref{supp:reward}), how each metric is computed (\cref{supp:metrics}), a comparison with the dense input, mesh decimation and remeshing (\cref{supp:dense}), additional mesh-quality metrics (\cref{supp:extra}), the VLM judge with its full prompt (\cref{supp:vlm}), the loss hyperparameters (\cref{supp:loss}) and further qualitative results (\cref{supp:figs}).

\section{Implementation details}
\label{supp:impl}
\mf~\cite{li2026meshflow} is used in geometry mode (dense-mesh condition only), in which the cross-attention (\texttt{attn2}) adapters receive exactly zero gradient and are left out, therefore we only apply lora on \texttt{attn1}. The reference model is the network with the adapters disabled. \Cref{tab:impl} lists the settings, and the loss hyperparameters are in \cref{tab:loss}.

\begin{table*}[t]
\centering
\footnotesize
\setlength{\tabcolsep}{4pt}
\begin{tabular}{@{}p{0.17\textwidth}p{0.79\textwidth}@{}}
\toprule
Item & Setting\\
\midrule
Base model & pretrained \mf~\cite{li2026meshflow}, geometry mode, only the LoRA adapters are trained\\
LoRA & rank 32, $\alpha{=}64$, weight 1, on self-attention \texttt{attn1} ($q,k,v$, out) and MLP (\texttt{fc1}, \texttt{fc2}), 96 modules\\
Sampler & released 28-step ODE, guidance 1.0, LoRA unmerged, for training data and evaluation\\
Renderer & 8 views (azimuth $0^\circ,45^\circ,\dots,315^\circ$, elevation $20^\circ$), perspective, field of view $40^\circ$, distance 1.6, $512^2$; input bounding-box frame (centre, 1/diagonal scale), 0.135\,s per mesh on the GPU\\
Hardware & one NVIDIA L40S GPU 48GB VRAM  (training, sampling, rendering, generation)\\
Training time & about 10 GPU-h in total for both rounds (sampling, scoring and updates)\\
\bottomrule
\end{tabular}
\caption{Overview of implementation details.}
\label{tab:impl}
\end{table*}

\paragraph{Rounds.}
We train two rounds. The first uses an earlier version of the reward and gating algorithm, without the silhouette and depth terms; the second, starting from the first-round model, uses the final reward and gating algorithm of \cref{sec:method}.

\section{Gating algorithm}
\label{supp:gating}
\Cref{tab:gates} lists every rule of \cref{sec:pairs} with its threshold. The winning sample $w$ is the best gate-passing sample of the group under $r$. The losing samples are every gate-failed sample, with $\omega=1$ whatever its reward, and the gate-passing samples that pass the clear-preference rule, with $\omega$ equal to their reward gap to $w$ divided by the largest such gap of the input. The fidelity and coverage filters apply to gate-passing losing samples, the face filter to both kinds, and the centre-offset filter, applied last, to all; $\omega$ is computed before the centre-offset filter and not rescaled after it. Inputs whose reward standard deviation is below 0.01 are skipped, as are inputs that keep no losing sample. Each update uses at most 3 losing samples per input (\cref{tab:loss}).

\begin{table*}[t]
\centering
\footnotesize
\setlength{\tabcolsep}{4pt}
\begin{tabular}{@{}p{0.20\textwidth}p{0.32\textwidth}p{0.44\textwidth}@{}}
\toprule
Rule & What it checks & Threshold\\
\midrule
\multicolumn{3}{@{}l}{\emph{Gates, relative to the samples of the same input}}\\
CD gate & fidelity & CD to $I$ $\le1.15\times$ group median\\
HD gate & fidelity, outliers & HD to $I$ $\le1.15\times$ group median\\
Flip gate & orientation & flipped-face fraction vs.\ $I$ $\le$ group median $+0.02$\\
Tiny-component gate & fragmentation & tiny components $\le1.5\times$ group median $+5$\\
Part-floor gate & no fusing of real parts & non-stray parts $\ge0.5\times$ group median\\
Decode gate & validity & decodes to a mesh\\
\midrule
\multicolumn{3}{@{}l}{\emph{Losing samples and pair filters (pair $w$ vs.\ $l$)}}\\
Clear preference & gate-passing $l$ clearly worse & $r(l)<r(w)-0.5\,\mathrm{std}(r)$ over the group\\
Fidelity filter & $w$ not worse in CD to $I$ & $\mathrm{CD}_I(w)\le1.02\,\mathrm{CD}_I(l)$\\
Coverage filter & $w$ not worse in uncovered area of $I$ & uncovered$_I(w)\le$ uncovered$_I(l)+0.002$\\
Face filter & no face shedding & faces$(w)\ge0.70\,$faces$(l)$\\
Centre-offset filter & $w$ not worse in bounding-box centre offset from $I$ & $\mathrm{off}(w)\le\mathrm{off}(l)+0.001$\\
\bottomrule
\end{tabular}
\caption{Gating criteria.}
\label{tab:gates}
\end{table*}

\section{Reward details}
\label{supp:reward}
\paragraph{Weights.}
We use the following weights $\lambda$ in \cref{eq:reward} (rounded to two decimals):
\begin{equation}
\begin{aligned}
r={}&0.2\,(t_{\text{sil}}+t_{\text{depth}})+0.15\,t_{\text{QS}}+0.1\,t_{\text{TS}}\\
&+0.05\,t_{\text{sliver}}+0.05\,t_{\text{val}}\\
&+0.1\,(t_{\text{frag-n}}+t_{\text{frag-stray}})+0.15\,\rdens .
\end{aligned}
\label{eq:reward-weights}
\end{equation}

\paragraph{Render and component scores.}
Silhouette IoU and depth L1 are computed per view over the union and the intersection of the silhouettes, the latter in units of the bounding-box diagonal, and averaged over the 8 views of the renderer in \cref{tab:impl}. Components are connected through shared edges after merging duplicate vertices: $n_{\text{tiny}}$ counts components with at most 4 faces, and $s_{\text{stray}}$ is the area share of components with less than $1\%$ of the total area that have a boundary edge or at most 4 faces.

\paragraph{Face-count target and band.}
$\rdens=\mathrm{clip}(1-|\log_2(F/F_t)|/2.5,0,1)$ with $F_t=\exp(4.60+0.471\ln F_{\text{ref8}})$, where $F_{\text{ref8}}$ is the median face count of 8 samples of the pretrained model on the input. The band width is $b{=}2.5$, so $\rdens$ is 1 at the target and falls to 0 at $2^{2.5}{\approx}5.7\times$ too few or too many faces.

\section{Metric computation}
\label{supp:metrics}
One harness scores every method; \cref{tab:main} and \cref{tab:dense} use the 300 test objects.
\paragraph{CD and HD ($\times100$).}
As in the \mf evaluation protocol: the artist mesh is scaled to the $[-0.5,0.5]$ unit cube and the output gets the same transform; 5,000 surface points are sampled on each, and the Chamfer and Hausdorff distances are computed with \texttt{point\_cloud\_utils}. The only change is a sampling seed fixed per object.
\paragraph{ECD ($\times100$).}
100k surface points are sampled per mesh. A point is an edge sample if the minimum of $|n_i\cdot n_j|$ over its (at most 24) neighbours within $\varepsilon=0.01$ is below 0.2. ECD~\cite{chen2021nmc} is the Chamfer distance between the edge samples of the output and of the artist mesh; it is undefined if either has fewer than 50 edge samples, so the means of \cref{tab:main} are over the 290--294 objects where it is defined.
\paragraph{Quad merge.}
Adjacent triangle pairs within $10^\circ$ of coplanar whose union is a convex quad with every corner in $[45^\circ,135^\circ]$ are candidates; they are taken greedily in decreasing order of their smallest corner, each triangle at most once. $\mathcal{Q}$ are the merged quads and $\mathcal{T}$ the remaining triangles. The same merge is used for QS, TS, the sliver term and every method.
\paragraph{QS (\cref{eq:qs}).}
Every quad $q$ has two nodes, one per pair of opposite sides. Two nodes are linked when their quads share a side that belongs to the side pairs of both nodes and that side is shared by exactly two quads. A strip $\pi$ is a connected component of this graph. It is a ring if it has at least 3 nodes and every node has degree 2, and $|\pi|$ counts its distinct quads. QS is the share of quads in $\mathcal{Q}$ that lie on a ring or on a strip with $|\pi|\ge4$, QS $=0$ if $\mathcal{Q}$ is empty.
\paragraph{TS (\cref{eq:ts}).}
TS~\cite{liu2025meshrft} $=0.4\,\tau_1+0.2\,\tau_2+0.3\,\tau_3+0.1\,\tau_4$. With corners $\theta_{q,k}$ and sides $L_{q,1},\dots,L_{q,4}$ (in cyclic order) of quad $q$: $\tau_1=|\mathcal{Q}|/\max(1,|\mathcal{Q}|+|\mathcal{T}|)$; $\tau_2=1-\tfrac{1}{|\mathcal{Q}|}\sum_q\sum_{k=1}^4|\theta_{q,k}-90^\circ|/360^\circ$; with $\rho_q=\max\bigl(\tfrac{\max(L_{q,1},L_{q,3})}{\min(L_{q,1},L_{q,3})},\tfrac{\max(L_{q,2},L_{q,4})}{\min(L_{q,2},L_{q,4})}\bigr)$, $\tau_3=\tfrac12\bigl(\tfrac{1}{|\mathcal{Q}|}\sum_q\rho_q\bigr)^{-1}+\tfrac{1}{2|\mathcal{Q}|}\sum_q\bar L_q/\max_k L_{q,k}$, $\bar L_q$ the mean side; and $\tau_4=\tfrac{1}{|\mathcal{Q}'|}\sum_{q\in\mathcal{Q}'}\bigl(1+\tfrac{1}{|N(q)|}\sum_{j\in N(q)}|\rho_q-\rho_j|\bigr)^{-1}$ over the quads $\mathcal{Q}'$ that share a side with at least one other quad, $N(q)$ those neighbours ($\tau_4=0$ if $\mathcal{Q}'$ is empty). If $\mathcal{Q}$ is empty, $\tau_2=\tau_3=\tau_4=0$.
\paragraph{Sliver and valence (\cref{eq:sliver}, \cref{eq:val}).}
$\alpha_f$ is the smallest corner of triangle $f$; $\deg(v)$ counts the edges of the triangle mesh at $v$, quad diagonals included.
\paragraph{DFF.}
DFF is the area share of faces that are on a boundary or non-manifold edge, self-intersect within a component, belong to a stray fragment ($<0.5\%$ of the area and open or with $\le4$ faces) or are degenerate (area below $10^{-9}$ of the total). Self-intersections are tested within each of the components that cover 95\% of the area (at most 50). FastMesh keeps every 3-clique of its predicted adjacency, so most of its area lies on non-manifold edges (DFF 0.990).
\paragraph{Frag.}
An output has introduced fragmentation if it has at least 5 more small components ($<1\%$ of the area) than the GT. Frag.\ is the share of such outputs. Fragmentation does not change monotonically: on \Vobj (3-seed mean component counts), the final model is closer to the GT component count than the pretrained \mf on 156 of 200 objects (5 ties), and on 35 objects the pretrained count is above the GT count and that of the final model below it.
\paragraph{Faces and s/mesh.}
Faces is the mean face count. s/mesh is the generation time per mesh recorded by the harness, averaged over the 300 test objects on one L40S GPU at batch 1. The classical methods of \cref{tab:dense} run on the CPU.

\section{Comparison with Classical baselines.}
\label{supp:dense}
Since many graphic artists use  classical remeshing algorithm on dense meshes, we compare our proposed model-based artist mesh generation performance with classical algorithm-based remeshing methods. QEM is pymeshlab quadric edge collapse (preserving boundary, face orientation and topology) to the face count of the MeshFlow. Isotropic remeshing is pymeshlab explicit isotropic remeshing with the edge length chosen so that face count is within MeshFlow's meshcount range (feature angle $60^\circ$, surface-distance check 1\%).

\Cref{tab:dense} compares them with \mf and \ours on the same objects. Our  model based methods shows better performance compared to algorithmic mesh generation in terms of QS and TS , which indicates that our proposed method is better in generating artist-preferred meshes. In algorithmic generation methods, they show extremely low fragmentation and DFF score, because algorithmic remeshing strongly keep the dense mesh structure so those are strictly following the shape of dense inputs.

\begin{table*}[t]
\centering
\footnotesize
\setlength{\tabcolsep}{0pt}
\begin{tabular*}{\textwidth}{@{\extracolsep{\fill}}lrrrrrrrrr@{}}
\toprule
Method & TS$\uparrow$ & QS$\uparrow$ & DFF$\downarrow$ & Frag.$\downarrow$ & CD$\downarrow$ & HD$\downarrow$ & ECD$\downarrow$ & Faces & s/mesh$\downarrow$\\
\midrule
\multicolumn{10}{@{}l}{\emph{Reference levels}}\\
Dense input & 0.655 & 0.711 & 0.000 & 0.083 & 2.299 & 5.615 & 3.711 & 95091 & --\\
\midrule
\multicolumn{10}{@{}l}{\emph{Decimation and remeshing of C1}}\\
QEM  & 0.527 & 0.321 & 0.024 & 0.083 & 2.343 & 6.132 & 4.902 & 611 & 3.45\\
Isotropic remeshing  & 0.492 & 0.169 & 0.014 & 0.083 & 2.471 & 7.845 & 7.484 & 1442 & 10.24\\
\midrule
\multicolumn{10}{@{}l}{\emph{Flow-based generation}}\\
\mf~\cite{li2026meshflow} & 0.575 & 0.466 & 0.206 & 0.743 & 2.760 & 6.169 & 8.149 & 609 & 2.17\\
\textbf{\ours} (\mf + ours) & 0.706 & 0.804 & 0.121 & 0.647 & 2.697 & 6.089 & 8.123 & 728 & 2.53\\
\bottomrule
\end{tabular*}
\caption{Dense input, mesh decimation and remeshing on the 300 test objects of \cref{tab:main} ( CD, HD, ECD $\times100$). The GT face budget comes from the artist mesh, which \ours never sees. Classical methods run on the CPU.}
\label{tab:dense}
\end{table*}

\section{Additional mesh-quality metrics}
\label{supp:extra}
QS, TS and DFF of \cref{tab:main} partly overlap with the terms of our reward. We therefore also measure three properties that no reward term computes: robustness to subdivision, non-manifold edges and automatic UV unwrapping.

\paragraph{Loop subdivision.}
We apply two levels of Loop subdivision~\cite{loop1987subdivision}. Edges with one or more than two incident faces are kept as creases~\cite{hoppe1994piecewise}. On the subdivided surface we measure the share of faces whose normal points against the normal of their original parent face (\emph{flip}), the share of faces selected by the self-intersection test of pymeshlab (\emph{self-int.}) and the share of faces that touch an extreme crease, an edge whose two face normals differ by more than $120^\circ$ (\emph{crease}). An artist's $90^\circ$ hard edge is not an extreme crease.

\paragraph{Non-manifold edges.}
\emph{NM edge} is the share of edges with more than two incident faces.

\paragraph{UV unwrapping.}
No method outputs UVs, so we unwrap every mesh with one fixed automatic unwrapper, xatlas 0.0.11 with its default chart and packing options. \emph{Charts} is the number of charts per 1,000 faces. Artist meshes have no ideal chart count, so we report the per-object distance $|x-x_{\text{artist}}|$ to the chart count of the artist mesh of the same object.

\begin{table}[t]
\centering
\footnotesize
\setlength{\tabcolsep}{3pt}
\begin{tabular}{@{}lrrrrr@{}}
\toprule
 & \multicolumn{3}{c}{Loop subdivision ($\times2$)} & & UV\\
\cmidrule(lr){2-4}\cmidrule(l){6-6}
Method & Flip$\downarrow$ & Self-int.$\downarrow$ & Crease$\downarrow$ & NM edge$\downarrow$ & Charts$^{\dagger}\downarrow$\\
\midrule
FastMesh~\cite{kim2026fastmesh} & 0.09 & 35.33 & 0.13 & 79.43 & 450.5\\
MeshAnything~\cite{chen2025meshanything} & 1.83 & 11.07 & 2.31 & 2.88 & 166.7\\
MeshAnything V2~\cite{chen2025meshanythingv2} & 1.34 & 12.35 & 2.45 & 4.63 & 121.6\\
Mesh Silksong~\cite{song2026silksong} & 0.77 & 2.52 & 0.66 & 1.88 & 63.6\\
\mf~\cite{li2026meshflow} & 1.10 & 3.43 & 0.39 & 11.04 & 84.3\\
\textbf{\ours} & 0.51 & 1.50 & 0.16 & 4.78 & 70.1\\
\bottomrule
\end{tabular}
\caption{Additional mesh-quality metrics on the 300 test objects of \cref{tab:main} ( means over objects). Flip, self-int., crease and NM edge are percentages of faces or edges. $^{\dagger}$: per-object distance to the chart count (per 1,000 faces) of the artist mesh.}
\label{tab:extra}
\end{table}

\paragraph{Results.}
\Cref{tab:extra} shows the results. After Loop subdivision, \ours has about half the flipped faces (0.51\% vs.\ 1.10\%), less than half the self-intersecting faces (1.50\% vs.\ 3.43\%) and less than half the faces at extreme creases (0.16\% vs.\ 0.39\%) of \mf. Its self-intersection share is the lowest of all methods, and its flip and crease shares are the lowest except for FastMesh, whose very dense meshes dilute per-face shares. \ours also has less than half the non-manifold edges of \mf (4.78\% vs.\ 11.04\%), far fewer than FastMesh and on par with MeshAnything V2, though more than MeshAnything and Mesh Silksong. On UV unwrapping, its chart count is closer to that of the artist mesh than that of \mf (70.1 vs.\ 84.3), a slight improvement, and closer than that of every other method except Mesh Silksong.

\section{VLM judge}
\label{supp:vlm}
\paragraph{Setup.}
The judge compares \ours with one other method at a time on 60 objects independent of the 300 test objects of \cref{tab:main}. Each call shows three images: the dense input, shaded only, and the two candidates, labeled only A and B, each as a $2\times2$ sheet of four fixed views (azimuth $30^\circ,120^\circ,210^\circ,300^\circ$, elevation $25^\circ$) with flat shading, a 1-px wireframe and back faces tinted red, all in the frame of the dense input. Every pair is judged in both orders. The per-object score is the share of the two orders in which \ours was chosen, and $W$ is its mean over objects, so an object judged differently in the two orders counts half.

Our judgement model is the state-of-the-art commercial VLM Claude Opus 5.5, since the VLM critic must mimic the expertised graphic artist. \Cref{fig:prompt} shows the full prompt.

\begin{figure*}[t]
\centering
\begin{minipage}{0.96\textwidth}
\begin{lstlisting}[basicstyle=\ttfamily\scriptsize, breaklines=true, columns=fullflexible, breakindent=0pt, upquote=true, frame=none]
===SYSTEM===
You are a senior 3D artist who reviews meshes for production use. You judge carefully and impartially, from what is visible in the images.
===INTRO===
Two candidate meshes, A and B, were made for the same input shape. You will see three images.
Image 1 shows the input: a dense reference mesh, shaded grey, from four viewpoints (2 x 2 grid).
Image 2 shows candidate mesh A and Image 3 shows candidate mesh B, from the same four viewpoints, with the same camera and scale as the reference. In the candidate images the surface is shaded light grey and every triangle edge is drawn as a thin dark line. Faces seen from their back side are tinted red: these are flipped (inverted) faces, or the inside of the surface seen through a hole or an open boundary.
===LABEL_REF===
Image 1: reference input (dense mesh, shaded only).
===LABEL_A===
Image 2: candidate mesh A.
===LABEL_B===
Image 3: candidate mesh B.
===QUESTION===
Which candidate is the better artist-quality mesh for this input? Weigh all of the following together:
1. Edge flow and topology: a clean, regular, purposeful edge layout that follows the shape, rather than a chaotic or noisy triangulation.
2. Triangle quality: well-shaped triangles, few thin slivers or needle triangles, and a sensible density for the shape.
3. Cleanliness: no stray floating fragments, no holes or missing pieces, no flipped faces.
4. Shape fidelity: faithful to the reference shape and its main parts.
The order in which the two candidates are shown is random and carries no information.
Reply with one JSON object and nothing else, exactly in this form:
{"rationale": "<at most 60 words comparing A and B>", "choice": "A"}
"choice" must be "A" or "B". You must pick one; a tie is not allowed.
\end{lstlisting}
\end{minipage}
\caption{The prompt of the VLM judge.  The lines \texttt{===SYSTEM===} to \texttt{===QUESTION===} separate the parts of one call: the system prompt, the introduction, the labels of the three images and the question. The images are sent as image blocks between the labels.}
\label{fig:prompt}
\end{figure*}

\section{Loss hyperparameters}
\label{supp:loss}
\Cref{tab:loss} lists the settings of the loss of \cref{eq:loss} of the final model.
\begin{table}[t]
\centering
\footnotesize
\setlength{\tabcolsep}{3pt}
\begin{tabular}{@{}p{0.30\columnwidth}p{0.66\columnwidth}@{}}
\toprule
Item & Setting\\
\midrule
$\beta$ & 2000\\
RPO weight $\gamma$ & 1.0\\
Weight $\omega$ & gate-passing losing sample: reward gap to $w$ over the largest gap of the input; gate-failed: 1 (\cref{supp:gating})\\
Losing samples & at most 3 per input per update, a uniform random subset redrawn every update\\
Noise level & one shared $(\epsilon,\sigma)$ per input, $\sigma=3u/(1{+}2u)$, $u\sim\mathcal{U}(0,1)$, clamped to $[10^{-3},1]$\\
Optimiser & AdamW, learning rate $10^{-4}$, weight decay $10^{-4}$, betas $(0.9,0.999)$; linear warm-up over 20 updates, then constant; gradient-norm clip 1.0\\
Batch & 4 inputs per update; per-input losses averaged\\
LoRA & rank 32, $\alpha{=}64$ (\cref{tab:impl})\\
Updates & 1,000 per round, each round on 1,000 new training inputs from the 2,000-object pool\\
Rounds & 2\\
\bottomrule
\end{tabular}
\caption{Loss and optimiser settings.}
\label{tab:loss}
\end{table}

\section{Additional qualitative results}
\label{supp:figs}
\begin{figure*}[t]
\centering
\includegraphics[width=\textwidth]{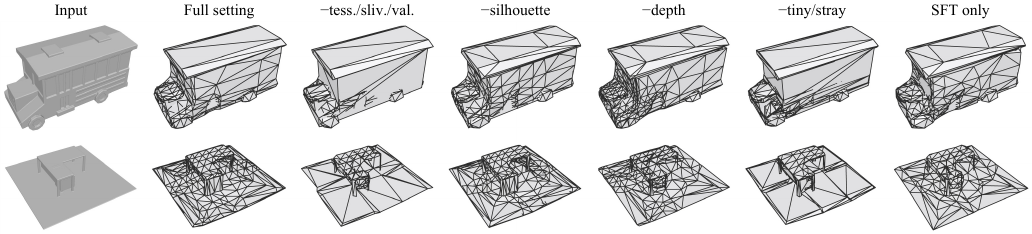}
\caption{Qualitative Ablation Study. We shoe the dense input, shaded, and the outputs of the full setting and five arms of \cref{tab:ablation_main}, with their mesh grids. }
\label{fig:supp-ablation}
\end{figure*}

\begin{figure*}[p]
\centering
\includegraphics[width=\textwidth]{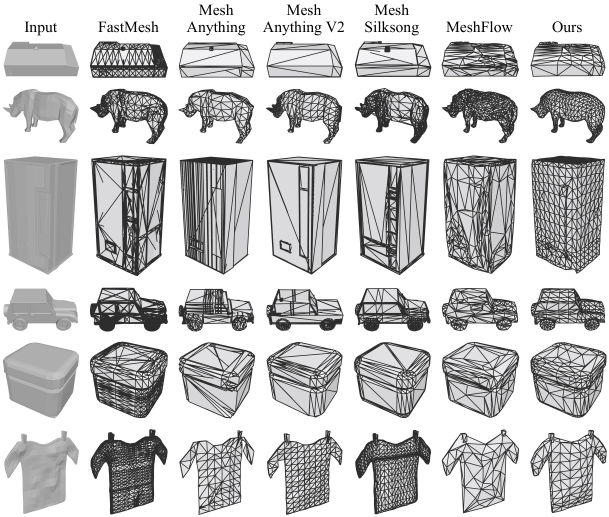}
\caption{Further comparisons on test objects: the dense input, shaded, and the output of each method with its mesh grid. }
\label{fig:supp-hero-a}
\end{figure*}

\begin{figure*}[p]
\centering
\includegraphics[width=\textwidth]{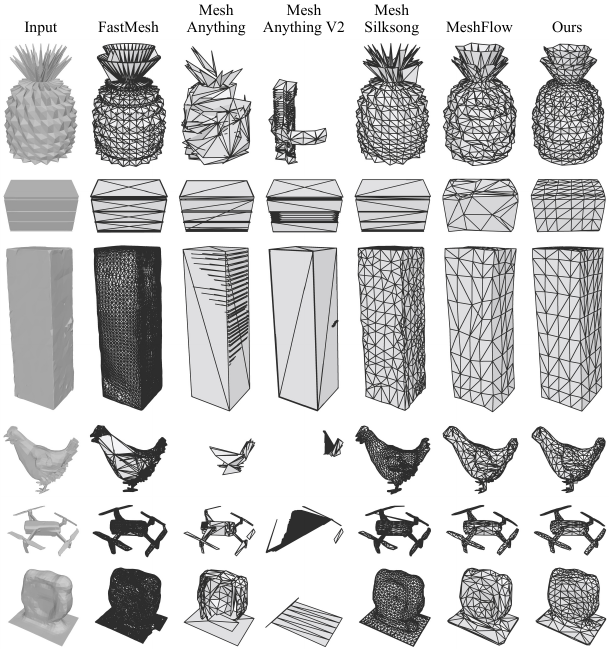}
\caption{Further comparisons on other test objects, as in \cref{fig:supp-hero-a}. }
\label{fig:supp-hero-b}
\end{figure*}

\Cref{fig:supp-ablation} shows outputs of the ablation study of \cref{tab:ablation_main} on two validation objects, and \cref{fig:supp-hero-a,fig:supp-hero-b} show further generated outputs from test objects. 

In \cref{fig:supp-ablation}, each row shows the dense input, the full setting and five arms: without the tessellation, sliver and valence terms, without silhouette IoU, without depth agreement, without the tiny and stray terms, and SFT only. The full setting covers the sides of the school bus and the floor slab under the desk with fairly even faces. Without the tessellation, sliver and valence terms, or without the tiny and stray terms, long thin triangles and fans appear on the same surfaces, and SFT only gives an irregular layout with many small triangles. Without depth agreement the layout is denser and less even; without silhouette IoU it is the closest to the full setting on these two objects. On average, every arm has a lower QS than the full setting in \cref{tab:ablation_main}, with the largest losses without the tessellation, sliver and valence terms and with SFT only.

\begin{figure}[t]
\centering
\includegraphics[width=\linewidth]{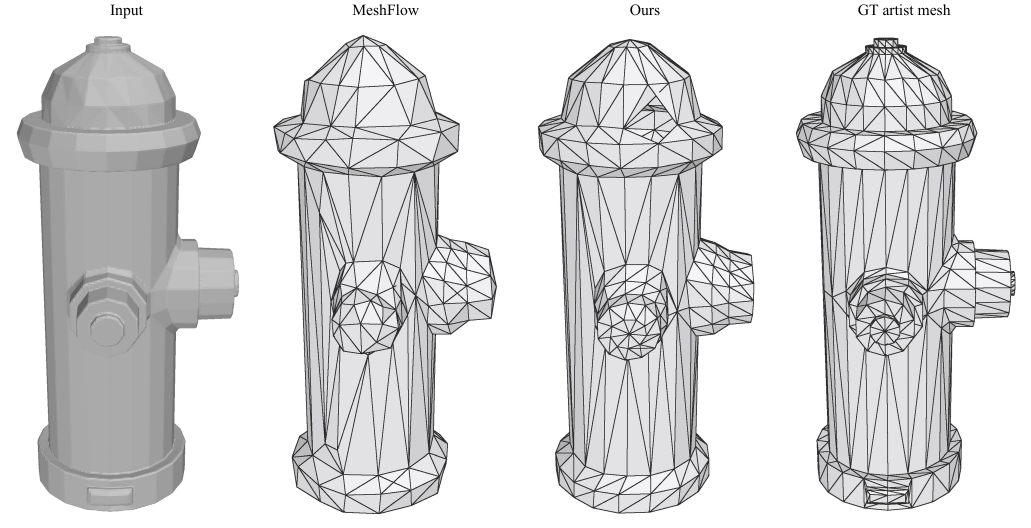}
\caption{\textbf{Failure cases of \ours.} Our model can improve the shape consistency and mesh structure, but still some of the defects can remain.}
\label{fig:supp-failure}
\end{figure}
\cref{fig:supp-failure} is failure case. Although our method can improve the mesh structure and shape consistency compared to baseline MeshFlow, our method can still shows unbalanced meshes or tiny components. Due to the property of flow based model, the output can be worse with different seeds. 

\end{document}